%% file: main.tex
\documentclass[10pt,twocolumn,letterpaper]{article}

\usepackage{iccv}              

\usepackage{dsfont} 
\usepackage{newtxtext,newtxmath}
\usepackage[T1]{fontenc}

\definecolor{iccvblue}{rgb}{0.21,0.49,0.74}
\usepackage[pagebackref,breaklinks,colorlinks,allcolors=iccvblue]{hyperref}
\usepackage[accsupp]{axessibility}  

\def\paperID{5567} 
\def\confName{ICCV}
\def\confYear{2025}

\title{Dynamic Reconstruction of Hand-Object Interaction with Distributed Force-aware Contact Representation}

\author{Zhenjun Yu\textsuperscript{1*}, Wenqiang Xu\textsuperscript{1*}, Pengfei Xie\textsuperscript{1}, Yutong Li\textsuperscript{1}, \\ Brian W. Anthony\textsuperscript{2}, Zhuorui Zhang\textsuperscript{2},  Cewu Lu\textsuperscript{1\S}\\ \textsuperscript{1}Shanghai Jiao Tong University, \textsuperscript{2}Massachusetts Institute of Technology \\
{\textsuperscript{1}\tt\small \{jeffson-yu, vinjohn, pf.xie, davidliyutong, lucewu\}@sjtu.edu.cn} \\
{\textsuperscript{2}\tt\small \{banthony, zhuoruiz\}@sjtu.edu.cn} \\ }

\begin{document}
\maketitle
\input{sec/0_abstract}    
\input{sec/1_intro}
\input{sec/2_related_works}
\input{sec/3_FO}
\input{sec/4_method}

\input{sec/5_dataset}
\input{sec/6_experiments}

\input{sec/7_conclusion}

\section*{Acknowledgement}
This work was supported by the Shanghai Commitee of Science and Technology, China (Grant No.24511103200) by the National Key Research and Development Project of China (No.2022ZD0160102), Shanghai Artificial Intelligence Laboratory, XPLORER PRIZE grants. This work has also been supported by the Shanghai Municipal Education Commission (No. 2024AIYB010) and the Fundamental Research Funds for the Central Universities (YG2025LC03).

{
    \small
    \bibliographystyle{ieeenat_fullname}
    \bibliography{main}
}

\end{document}


\maketitle

\section{Supplementary Video}\label{sec:video}
In the supplementary material, we provide a video that visually demonstrates several key information in our paper. The video is organized as follows:

\begin{itemize}
    \item \textbf{0:00--0:17}: Qualitative results on DexYCB Dataset for comparing our method and gSDF \cite{gsdf}.
    \item \textbf{0:17--0:30}: Qualitative results on HOT Dataset for comparing our method with CPF \cite{cpf}, TOCH \cite{toch} and ViTaM \cite{vitam}.
    \item \textbf{0:30--0:42}: Hardware setup for our real-world experiments.
    \item \textbf{0:42--1:00}: Real-world experiment results on two stuffed toys.
\end{itemize}

The video is intended to complement our demonstration in the main paper by providing dynamic illustrations of our experimental results.

\section{Architecture of Flow Prediction Module}\label{sec:flow_arch}
We introduce the detailed architecture of our proposed flow prediction module $\mathcal{F}_f(\cdot)$. Assuming at frame $t$, the per-point features $F_t$, $F_{t-1}$ is extracted from $\mathcal{P}_t$ and $\mathcal{P}_{t-1}$, the flow prediction module takes the two set of point clouds and their feature to predict the point cloud flow between two frames:
\begin{equation} \label{eq:corr}
    f_{t-1 \rightarrow t} = \mathcal{F}_f(F_t, F_{t-1}, \mathcal{P}_t, \mathcal{P}_{t-1}).
\end{equation}
We first perform a Cartesian product of the two extracted features, yielding a tensor of size $N \times N \times 2d$. This tensor is fed into a 3-layer MLP to obtain $\mathcal{C}_v$, which is then used in two ways. First, it passes through a 2D convolutional layer for downsampling to obtain $p_c \in \mathbb{R}^{N \times N}$, representing the matching probability of each point between two frames. Second, $\mathcal{C}_v$ is sent through a softmax function and a one-layer MLP to downsample, resulting in $\mathcal{C}_c \in \mathbb{R}^{N}$, indicating whether the points in the first frame are matched in the second frame. Thus, the final matching probability matrix $p_m \in \mathbb{R}^{N \times N}$, which describes the correspondence likelihood between the two point sets, can be computed as:

\begin{equation}
    p_m = p_c \times \mathcal{C}_c
\end{equation}

After estimating the matching probability, we compute the disparity of two point cloud sets $\mathcal{D} \in \mathbb{R}^{N \times N \times 3}$, with $\mathcal{D}_{ij} = \mathcal{P}_t^i - \mathcal{P}_{t-1}^j$, and concatenate the disparity with $p_m$. The concatenated tensor is then fed into four 2D convolutional layers with batch normalization and a softmax function to obtain the disparity feature $F_d \in \mathbb{R}^{N \times d^{\prime}}$. Finally, we use PointNet++ to regress the flow $f_{t-1 \rightarrow t} \in \mathbb{R}^{N \times 3}$.

In practice, the point cloud feature size is $d = 128$, and the disparity feature size is $d^\prime = 64$.

\section{Flow Prediction Result}\label{sec:app_flow}
To validate the accuracy of our proposed flow prediction module, we report the Chamfer distance error on the DexYCB and HOT datasets in Tab. \ref{tab:flow_pred}. The relatively low Chamfer distance demonstrates the efficacy of our network. The slightly better results on the DexYCB dataset are likely due to the larger hand-object movements and more challenging object deformations in the HOT dataset.

\begin{figure}[th!]
    \centering
    \includegraphics[width=1\linewidth]{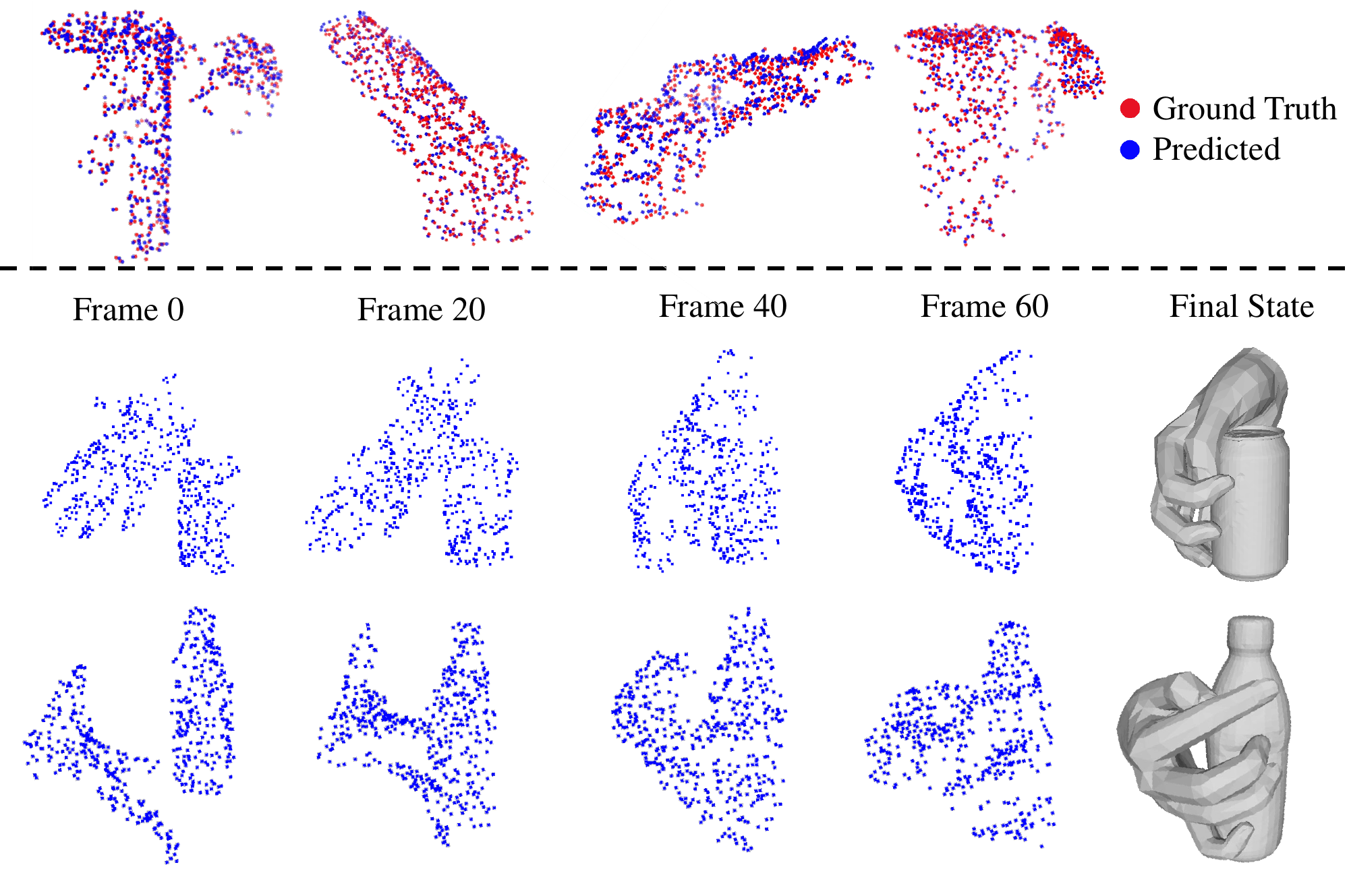}
    \caption{Qualitative results on flow prediction using the Flow Prediction Module.}
    \label{fig:qual_flow}
    \vspace{-0.3cm}
\end{figure}

We also present some qualitative results in Fig. \ref{fig:qual_flow}. The top section compares our predicted flow added to the last frame's point cloud with the ground truth of the current frame's point cloud. The near overlap of the two point clouds indicates high prediction accuracy. The bottom section shows a sequence of our estimation results, illustrating our method's ability to track hand movements and object deformations in whole sequences.

\begin{table}[!th] 
    \center
    \begin{tabular}{c|c|c|c|c}
        \toprule
        Dataset & \multicolumn{2}{c|}{DexYCB} & \multicolumn{2}{c}{HOT}  \\ \hline
        
        Category & Box & Bottle & Sponge & Plasticine \\

        \midrule
        CD(mm)$\downarrow$ & 8.7 & 9.1 & 10.3 & 12.2  \\ 

        \bottomrule

    \end{tabular}
    \caption{Quantitative results for flow predictions in visual dynamic tracking net.}
    \label{tab:flow_pred}
\end{table}

\section{Ablation Study, Extended}

\subsection{Importance of flow prediction module}\label{sec:app_flow_prediction}
To demonstrate the importance of our flow prediction module, we conducted an experiment by directly fusing the visual features extracted from the input point clouds using the transformer fusion layer. The quantitative results are shown in Tab. \ref{tab:corr_abl}. Introducing flow prediction module significantly improves all scores, validating our feature fusion approach that incorporates the 3D static information of the current frame and the correspondence feature from the flow.

\begin{table}[!th] 
    \center
    \begin{tabular}{c|c|c|c}
        \toprule
        Metrics & IoU$\uparrow $ & CD$\downarrow$ & MPJPE$\downarrow$ \\ 
        \midrule
        Dataset & \multicolumn{3}{c}{DexYCB} \\
        \midrule

        w/o Flow Pred. & 84.3 & 15.7 & 17.1  \\ 
        w. Flow Pred. & \textbf{90.1} & \textbf{9.6} & \textbf{13.2}  \\ 

        \midrule

        Dataset & \multicolumn{3}{c}{HOT} \\
        \midrule
        w/o Flow Pred. & 64.7 & 20.2 & 16.3  \\ 
        w. Flow Pred. & \textbf{80.5} & \textbf{10.9} & \textbf{13.6}  \\ 

        \bottomrule

    \end{tabular}
    \caption{Quantitative results on DexYCB and HOT Dataset for whether using the flow prediction module. "Flow Pred." stands for flow prediction module, "w/o" indicates without.}
    \label{tab:corr_abl}
    \vspace{-0.3cm}
\end{table}

\subsection{Different point pair establishment strategies}\label{sec:app_point_pair}
This section discusses the influence of two point pair establishment strategies: using \textbf{keypoints} or \textbf{all-hand vertices}. When considering all hand vertices, we establish point pairs between them and nearby object vertices, treating the force exerted by the hand as the reading from the nearest sensor. The quantitative results of these two methods are shown in Tab. \ref{tab:key_sensor}. While penetration depth improves slightly, both contact IoU and MPJPE decrease. This is likely because considering all sensors leads to conflicting optimization directions for the same joint, as sensors within the same regions may experience different contact situations. Additionally, the iteration time increases about tenfold compared to our setting.

\begin{table}
    \center
    \begin{tabular}{c|c|c|c|c}
        \toprule
        Metrics & MPJPE$\downarrow$ & PD$\downarrow$ & CIoU$\uparrow$ & Iter. Time(s)$\downarrow$ \\ 
        \midrule

        Keypoint & \textbf{11.3} & 7.3 & \textbf{40.3} & \textbf{3.5 $\pm$ 0.5} \\ 
        All Verts & 14.5 & \textbf{6.9} & 25.6 & 37 $\pm$ 3 \\ 

        \bottomrule

    \end{tabular}
    \caption{Quantitative results on evaluating point pair establishment on key points and on all-hand vertices.}
    \label{tab:key_sensor}
    \vspace{-0.5cm}
\end{table}

\subsection{Tactile feature fusion in visual dynamic tracking net.}
To assess the impact of introducing distributed tactile arrays in visual dynamic tracking net, we first use a 3-layer MLP to encode the tactile features of each region. By estimating hand pose, we fuse these regional features to the sample points, adding the encoded tactile feature to the point-wise feature of each sample position. We train visual dynamic tracking net with fused tactile information on our HOT dataset, and the results are shown in Tab. \ref{tab:abl_tac}. Quantitative results show no significant improvements, likely because the tactile data are much more sparse than the visual inputs, causing feature misalignment. Therefore, we implement DF-Field to convert force readings into contact states for hand-pose optimization. 

\begin{table}[ht]
    \center
    \begin{tabular}{c|c|c}
        \toprule
        Metrics & IoU$\uparrow$ & CD$\downarrow$ \\ 
        \midrule

        w/o Tactile Fusion & 81.0 & \textbf{10.9}  \\ 
        w. Tactile Fusion & \textbf{81.3} & 11.5  \\

        \bottomrule

    \end{tabular}
    \caption{Quantitative results for whether or not fusing tactile information in visual dynamic tracking net. "w/o" indicates without.}
    \label{tab:abl_tac}
    \vspace{-0.3cm}
\end{table}

\subsection{Effectiveness of Energy and Loss Terms for Force-aware Optimization}

To better understand the contribution of each component in our force-aware optimization, we conduct ablation studies on the HOT dataset, evaluating the effects of removing barrier energy $B_{ij}$, relative potential energy $E_{ij}$, and loss terms $\mathcal{L}_r$ and $\mathcal{L}_o$ from our framework.

The results are reported in Table \ref{tab:ablation_energy}, demonstrating the importance of each term in maintaining physical plausibility and reconstruction accuracy. Removing the Barrier Function leads to significant interpenetration between the hand and object, and weakens contact quality due to the unbalanced attraction from the Relative Potential Energy. Conversely, eliminating the Relative Potential Energy causes the Barrier Function to over-separate the hand from the object, resulting in poor contact recovery despite low penetration.

Without the regularization term $\mathcal{L}_r$, fingers may exhibit unnatural poses, leading to increased joint error. Omitting $\mathcal{L}_o$ causes the reconstructed hand pose to deviate notably from the initial prediction, as evidenced by a large increase in MPJPE (20.2 mm).

These results validate the necessity of combining all proposed energy and loss terms for physically consistent and accurate dynamic hand-object reconstruction.

\begin{table}[ht]
    \centering
    \begin{tabular}{c|c|c|c}
    \toprule
    Config. & MPJPE(mm)$\downarrow$ & PD(mm)$\downarrow$ & CIoU(\%)$\uparrow$ \\
    \midrule
    Full Opt. & \textbf{11.3} & 7.3 & \textbf{40.3}  \\
    \midrule
    w/o $B_{ij}$ & 13.8 & 14.5 & 25.0  \\
    w/o $E_{ij}$ & 17.2 & \textbf{2.5} & 12.0  \\
    \midrule
    w/o $\mathcal{L}_r$ & 15.7 & 8.4 & 33.2  \\
    w/o $\mathcal{L}_o$ & 20.2 & 8.1 & 35.1  \\
    \bottomrule
    \end{tabular}
    \caption{Ablation study of energy and loss terms on the HOT dataset. }
    \label{tab:ablation_energy}
\end{table}


\section{Limitation}

Our current approach mainly focuses on single-hand interaction scenarios and leverages depth input for reliable geometry observation rather than RGB settings. While effective, the method assumes a reasonably reconstructed object mesh; in rare cases where object geometry is significantly degraded, force-aware optimization may be affected. Extending the framework to handle multi-hand interactions, further improving object reconstruction robustness, and refinement for object meshes based on force optimization, remain valuable directions for future work.

{
    \small
    \bibliographystyle{ieeenat_fullname}
    \bibliography{main}
}

%% file: sec/0_abstract.tex

\begin{abstract}
We present ViTaM-D, a novel visual-tactile framework for reconstructing dynamic hand-object interaction with distributed tactile sensing to enhance contact modeling. Existing methods, relying solely on visual inputs, often fail to capture occluded interactions and object deformation. To address this, we introduce DF-Field, a distributed force-aware contact representation leveraging kinetic and potential energy in hand-object interactions. 
ViTaM-D first reconstructs interactions using a visual network with contact constraint, then refines contact details through force-aware optimization, improving object deformation modeling. To evaluate deformable object reconstruction, we introduce the HOT dataset, featuring 600 hand-object interaction sequences in a high-precision simulation environment. 
Experiments on DexYCB and HOT datasets show that ViTaM-D outperforms state-of-the-art methods in reconstruction accuracy for both rigid and deformable objects. DF-Field also proves more effective in refining hand poses and enhancing contact modeling than previous refinement methods. The code, models, and datasets are available at \url{https://sites.google.com/view/vitam-d/}.
\end{abstract}

\renewcommand{\thefootnote}{}
\footnotetext{\noindent * indicates equal contributions. \S Cewu Lu is the corresponding author.}

%% file: sec/1_intro.tex
\section{Introduction}
Humans manipulate objects using both visual and tactile feedback intuitively. Modeling and reconstructing this hand-object interaction can empower many downstream tasks, including VR/AR, robotic imitation learning, and human behavior understanding. Existing methods for dynamic hand-object interaction reconstruction \cite{cpf,cpf_pami,hotrack,vo_ho1,vo_ho2} primarily rely on vision to recover the global geometry and estimate hand-object poses. However, these methods struggle to capture the occluded contact details, particularly the object deformation at the contact interface.

\begin{figure}[h!]
    \centering
    \includegraphics[width=1\linewidth]{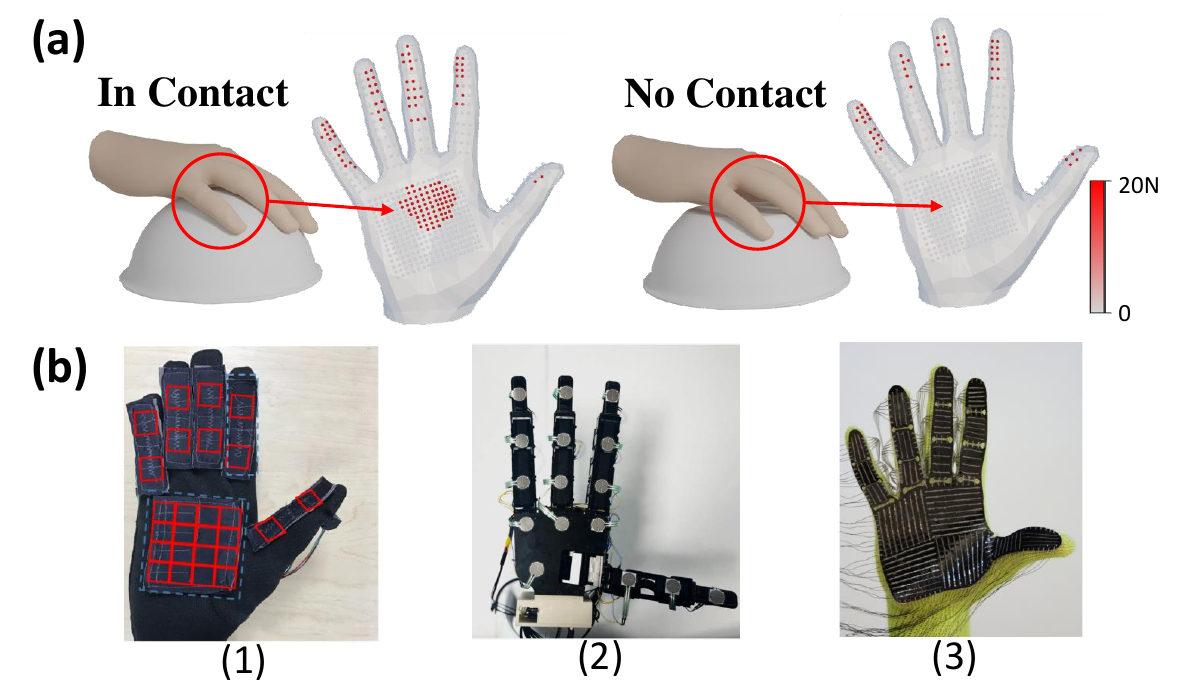}
    \caption{(a): The relationship between tactile information and contact geometry. Grasping different bowls with the same hand poses shows that distributed tactile arrays can capture occlusion contacts and object states. (b): Different types of distributed tactile sensors proposed in previous works (1). \cite{glove1}, (2). \cite{glove2}, (3). \cite{glove_nature}.}
    \label{fig:demo}
    \vspace{-0.5cm}
\end{figure}

Recent advances in tactile sensing \cite{glove_nature,glove1,gelsight,mc_tac} inspire the development of visual-tactile approaches for hand-object reconstruction tasks \cite{vtaco,vt_recon1,vt_recon2} to leverage fine-grained perceptual signals from contact regions. (Fig. \ref{fig:demo}.a). Among various tactile sensor types, distributed tactile sensors (Fig. \ref{fig:demo}.b) \cite{glove1,glove2,glove_nature}, in form factors of wearable gloves or attachable sensor arrays, show great promise in human manipulation data collection comparing to other counterparts \cite{gelslim,gelsight,digit,mc_tac}.  A previous work, ViTaM \cite{vitam} proposed a novel high-density tactile glove system that is capable of generating high-accuracy force data. Meanwhile, ViTaM utilized a transformer module to directly fuse tactile sensory data with visual perception for object reconstruction, which ignores the unbalanced information between forces and point cloud features. Therefore, how to effectively integrate distributed tactile sensors and state-of-the-art visual perception tools for refined hand–object reconstruction remains underexplored.

Since distributed tactile sensors can conform closely to the hand while maintaining the hand's geometry and natural dexterity during manipulation, we can use an off-the-shelf visual approach to reconstruct the hand-object state and integrate the tactile information to refine the results. In light of this, we propose a novel \textbf{D}istributed \textbf{F}orce-aware contact representation, \textbf{DF-Field}, and a \textbf{Vi}sual-\textbf{Ta}ctile \textbf{M}anipulation reconstruction framework with \textbf{D}istributed tactile sensing, \textbf{ViTaM-D}. DF-Field models the contact by considering both kinetic and potential energy in hand-object interaction. This approach enables accurate modeling of object deformation.
Using DF-Field, ViTaM-D initially reconstructs the hand-object interaction with visual observations and contact state constraints by the proposed visual dynamic tracking network, and then refines the contact details with DF-Field via a force-aware hand-pose optimization process. The proposed framework leverages advances in visual-only hand-object reconstruction methods \cite{cpf_pami,hotrack,vo_ho1,vo_ho2} while seamlessly integrating the tactile information into an existing motion capture or estimation system to enhance the fidelity of contact modeling and overall reconstruction quality.

To build and train such a visual-tactile framework, large-scale datasets that capture diverse hand-object interactions with precise tactile readings are required, especially for deformable objects. However, existing datasets for hand-object interactions \cite{dexycb,oakink,arctic} only cover rigid or articulated object manipulation and generally lack accurate tactile data. 
Therefore, alongside using the public dataset (\eg DexYCB \cite{dexycb}) to benchmark on \textit{rigid} objects with common baselines \cite{gsdf, hotrack}, we also introduced a new dataset, \textbf{HOT dataset}, to fully benchmark our method on \textit{deformable} object reconstruction. This dataset is built with ZeMa \cite{zema}, a high-precision physics-based simulation environment that supports penetration-free frictional contact modeling with finite element method (FEM) to model the object deformation. The HOT dataset contains 600 sequences of hand-object interaction, with 30 deformable objects from 5 different categories and 8 camera views for each sequence.

To evaluate the method, we compare our proposed ViTaM-D with previous state-of-the-art methods gSDF \cite{gsdf}, HOTrack \cite{hotrack} on DexYCB, and ViTaM\cite{vitam} on our proposed HOT dataset. Results show that our method significantly improves the reconstruction performance compared to previous works. We successfully demonstrated the superiority of our method in tracking both rigid and deformable objects. Besides, the effectiveness of the force-aware optimization has been validated in refining hand poses to eliminate penetrations and suboptimal contact states.

Our contributions are summarized as follows:

(1) A visual-tactile learning framework, ViTaM-D. It contains a visual dynamic tracking network for reconstructing hand-object interactions, and a force-aware optimization process, which integrates the tactile information into reconstruction refinement based on a novel distributed force-aware contact representation, DF-Field.

(2) A new dataset, HOT. It contains 600 RGB-D manipulation sequences on 30 deformable objects from 5 categories with penetration-free hand-object poses and accurate tactile information.

%% file: sec/2_related_works.tex
\section{Related Work}
Hand-object reconstruction has been richly studied because it tries to recover the full details of hand-object interaction, showing potential applicability for downstream tasks.

Research in this direction starts with static reconstruction. Earlier works have predominantly relied on RGB inputs to estimate hand pose and object geometry. Hasson et al. \cite{hasson2019learning} presented a method that learns hand-object interaction using synthetic RGB images with the grasp poses planned by GraspIt! \cite{graspit}. Doosti et al. \cite{doosti2020hope} proposed a graph-based network for hand-object pose estimation. Cao et al. \cite{cao2021reconstructing} introduced a method that operates in-the-wild, estimating the hand pose first and optimizing the solution using a contact representation. The optimization involves push and pull terms to handle the contact, which is purely empirical. Similarly, CPF \cite{cpf, cpf_pami} employs a contact representation using a spring-mass system, adopting an empirical approach to refine the hand-object interaction. ArtiBoost \cite{artiboost} further enhanced the performance of static hand-object reconstruction through data augmentation techniques. AlignSDF \cite{alignsdf} leveraged RGB inputs to estimate hand pose and combined point cloud data for object geometry, using SDF-based decoders for both hand and object reconstruction.

Later, hand-object reconstruction in the dynamic setting draws increasing attention since manipulation is naturally dynamic. Approaches have advanced by incorporating temporal information and using more complex representations. Tekin et al. \cite{tekin2019h+} adopted RNN for temporal feature fusion, utilizing egocentric RGB video to capture dynamic hand-object interaction and pose estimation. Hasson et al. \cite{hasson2020leveraging} improved dynamic reconstruction by enforcing photometric consistency constraints. Ye et al. \cite{ye2023diffusion} introduced a diffusion model that guides dynamic hand-object reconstruction via score distillation sampling and differentiable rendering techniques. Recent advances by Fan et al. \cite{fan2024hold} focused on refining pose estimation using SDF-based representations and volumetric rendering, improving the overall dynamic reconstruction accuracy. Furthermore, gSDF \cite{gsdf} employed transformer architectures and SDF representations to model complex hand-object interactions dynamically.

These works focus more on visual-only inputs. However, due to occlusion between the hand and object during the interaction, visual perception usually lacks information near the contact areas, and such information cannot necessarily be mitigated by cross-frame feature fusion. Therefore, tactile perception comes to supplement the near-contact information. Zhang et al. \cite{zhang2021dynamic} utilized a tactile glove \cite{glove_nature} to track object movements. However, it focuses more on the dynamic object trajectory rather than contact geometry. Works such as \cite{vt_recon1, vt_recon2} employed optical tactile sensors, trained the model by synthetic data, and applied to rigid object geometry reconstruction. Later, VTacO \cite{vtaco} extended this line of research by using optical tactile sensors to capture object geometry, including deformation, providing a more comprehensive representation of hand-object interactions. Jiang et al. \cite{vitam} proposed a visual-tactile recording and tracking system ViTaM, which contains a wearable tactile glove along with a visual-tactile reconstruction algorithm. However, it simply used neural networks to fuse tactile and visual data and did not consider the information mismatch between them.  Unlike these works, the proposed ViTaM-D inherits the merits from both worlds: it leverages the advanced techniques of visual reconstruction and incorporates the distributed tactile readings to enhance the local contact details.

%% file: sec/3_FO.tex
\section{Force-aware Contact Representation}\label{sec:cpf}
During manipulation, forces are reciprocal, so the force exerted by the hand affects the state of both the hand and the object mutually. To fully capture dynamic contact behaviors, the representation must encode both contact locations and forces. We introduce a distributed force-aware contact representation, \textbf{DF-Field} (Sec. \ref{sec:fc_field}), and apply it using distributed tactile sensors (Sec. \ref{sec:fc_tactile}).

\subsection{DF-Field Representation}\label{sec:fc_field}

Without loss of generality, we take the object-centric perspective to describe a hand-object interaction process, where the object is assumed to be fixed in the origin point, and the hand moves around the object. In this way, by ignoring the gravitational potential energy, the rest of the contact dynamics is the kinetic energy and deformation potential energy of the hand and object, which are described by a \textbf{Relative Potential Energy}. Besides, to prevent penetration between the hand and object contact surface, we assume a virtual \textbf{Barrier Energy}. With an established point pair $i,j$ of object vertex $\mathcal{V}_i^o$ and hand vertex $\mathcal{V}_j^h$, and the Euclidean distance of the point pair $l_{ij} = \left\lVert \mathcal{V}_i^o - \mathcal{V}_j^h \right\rVert_2$, we define the two energy terms as follows.

\textbf{Relative Potential Energy.} A relative potential energy that describes both the object deformation and the hand movements:
\begin{equation}
    E_{ij} = \kappa l_{ij}^2,
\end{equation}
where $\kappa$ is a parameter representing the hand interacting with the object vertices. If $\kappa > 0$, the hand and object are in contact. Thus, the distance $l_{ij}$ and the relative energy will be close to $0$, indicating that the contact's relative potential energy is satisfied.

\textbf{Barrier Energy.} Given a certain threshold distance $\hat{l}$, the barrier energy is:
\begin{equation}
    B_{ij} = \left\{\begin{array}{ll}
        - e^{-\kappa} (l_{ij} -\hat{l})^{2} \log \left(\frac{l_{ij} }{\hat{l}}\right), & 0<l_{ij} <\hat{l} \\
        0 & l_{ij} \geq \hat{l}
        \end{array}\right.
\end{equation}
Adopted from \cite{IPC}, the barrier energy aims to push away the point pair when $\kappa$ is low, thus avoiding penetration issues between the hand and object. The function is defined in this way not only to ensure repulsion becomes smaller when the distance is larger but also to remain smooth for optimization.

\textbf{Overall Energy.} A proper contact is met when both energy terms approach 0:
\begin{equation}\label{eq:energy}
    E = \sum \limits_{i} \sum \limits_{j} (E_{ij} + B_{ij}).
\end{equation}

To note, though $\kappa$ is strongly correlated to the tactile readings, DF-Field can also be tactile-independent, as long as we empirically set the exerted force instead of physics-based. In this way, DF-Field can work with visual-only methods. The details are discussed in Sec. \ref{sec:ablation}.

\begin{figure}
    \centering
    \includegraphics[width=0.9\linewidth]{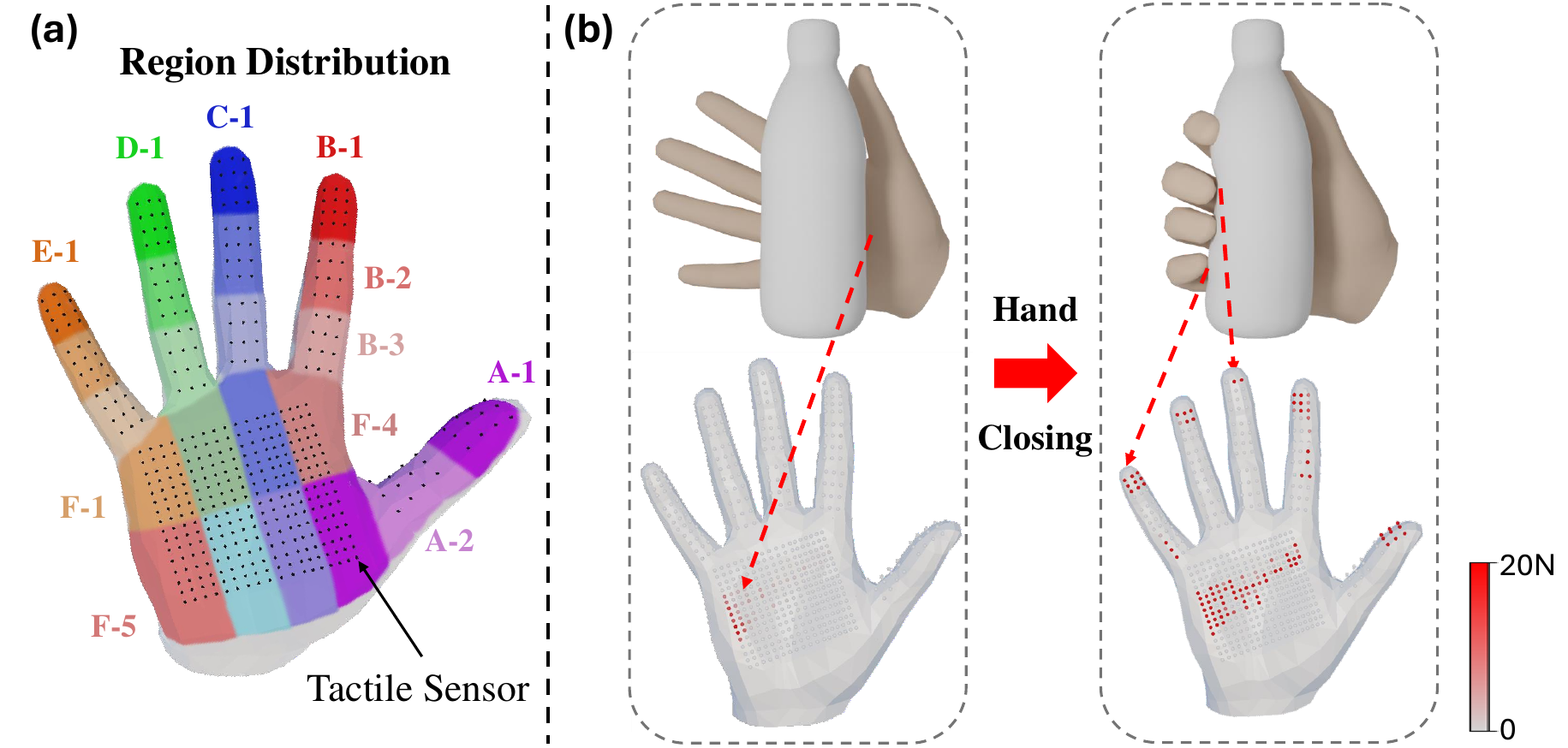}
    \caption{(a): The 22 regions and a typical distributed tactile sensor layout \cite{glove_nature}. (b): An example of data collection in the simulation environment, generating depth images and tactile sensor readings.}
    \label{fig:sim}
    \vspace{-0.5cm}
\end{figure}

\subsection{DF-Field with Distributed Tactile Sensors}\label{sec:fc_tactile}
As aforementioned, the distributed tactile sensors generally conform to the hand, and different tactile sensors may have different layout configurations. Thus, to adapt to different tactile sensors, we take the hand-centric perspective by first dividing the hand into 22 regions (as shown in the left of Fig. \ref{fig:sim}.a, 2 areas for the thumb, 3 for other fingers, and 8 for the palm), and assign the tactile sensors to the corresponding region. In each region, we define the center as the hand keypoint, resulting in $\mathcal{K}^h \in \mathbb{R}^{22 \times 3}$. 

For optimization, we connect only the hand keypoints to object vertices, reducing computational load while enhancing regional interaction information. The force in each region is calculated by averaging the tactile readings $\mathcal{M}^j$. By the definition of the energy, we can obtain that $\kappa$ is the difference of the exerted force over the distance $l_{ij}$. In practice, we approximate it by dividing the force by the distance:
\begin{equation}
    \kappa_{ij} \sim \frac{\overline{\mathcal{M}^j}}{l_{ij}}.
\end{equation}
We discuss the difference between using keypoints or all-hand vertices for optimization in the supplementary file.

%% file: sec/4_method.tex
\section{ViTaM-D Method}\label{sec:method}

\begin{figure*}[th!]
    \centering
    \includegraphics[width=1\linewidth]{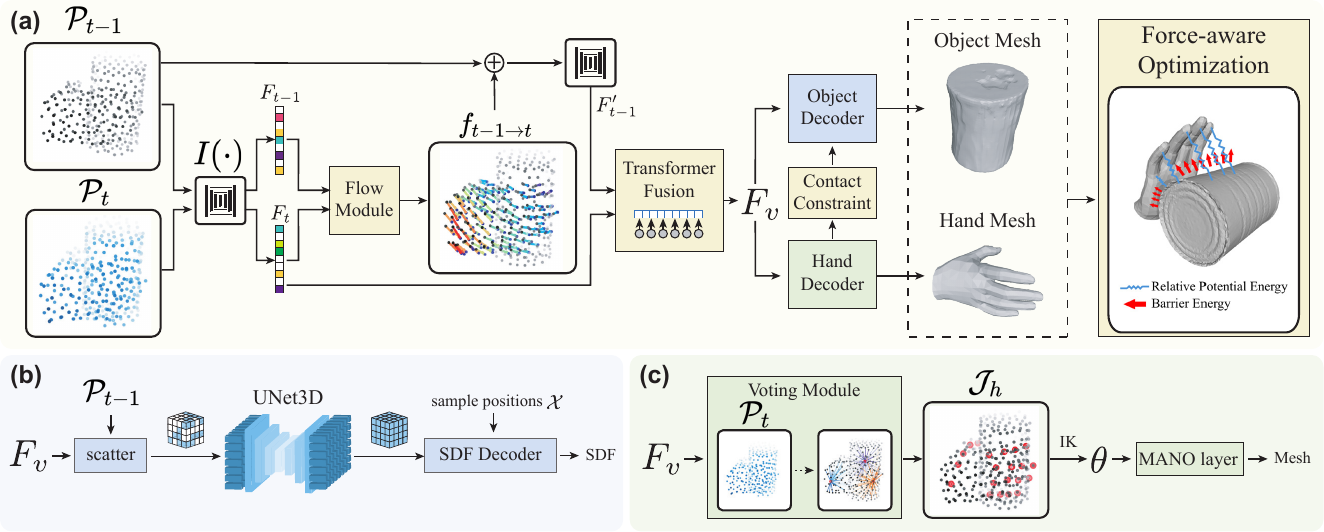}
    \caption{\textbf{ViTaM-D} pipeline. (a): The overview of our pipeline, including the Flow Prediction module for visual feature extraction and flow estimation, Hand and Object Decoders, the Contact Constraint, and Force-aware hand-pose Optimization.  (b): The Object Decoder to reconstruct the object mesh. (c): The Hand Decoder for estimating hand parameters and reconstructing based on the MANO model. }
    \label{fig:pipeline}
    \vspace{-0.3cm}
\end{figure*}

\subsection{Overview}
We address the problem of 4D tracking and dynamic reconstruction of hand-object interactions from a visual-tactile perspective. Based on DF-Field, we propose a novel method named \textbf{ViTaM-D}, which consists of two stages: visual dynamic tracking and force-aware optimization. We assume that the visual input consists of a live stream of 3D point clouds, denoted as $\{\mathcal{P}_t \in \mathbb{R}^{N \times 3} \}_{t=1}^n$, representing hand-object interactions derived from single-view depth images, where $N$ is the number of input point, $n$ is the number of frames. The tactile input $\{\mathcal{M}_t\}_{t=1}^n$ contains distributed tactile sensor readings for each frame.

In the first stage, ViTaM-D uses visual input to establish flow features and dynamically reconstruct the hand and object. Contact information is used to constrain the reconstruction to capture global hand-object geometry. In the second stage, tactile information is incorporated to improve contact accuracy and reduce penetration issues based on our proposed DF-Field. The overall pipeline is given in Fig. \ref{fig:pipeline}.

\subsection{Visual dynamic tracking of hand and object}

In the first stage, to introduce correspondence feature from the previous frame, the visual dynamic tracking net predicts the point cloud flow from the last to the current frame using a \textit{flow prediction module}, and extracts a fused visual feature containing both correspondence feature and static information. This feature is forwarded into an \textit{object decoder} and a \textit{hand decoder} to reconstruct the object and hand geometry, respectively. To improve global geometry accuracy, we introduce a novel \textit{contact constraint} that reasons about object geometry based on hand reconstruction. Objects are modeled using signed distance fields (SDF), while the hand is represented using the MANO model \cite{mano}.

\textbf{Flow Prediction Module.}\label{para:corr}
At frame $t$, we first extract the per-point features $F_t, F_{t-1} \in \mathbb{R}^{N \times d}$ from $\mathcal{P}_t$ and $\mathcal{P}_{t-1}$ using a backbone network $I(\cdot)$. In practice, we adopt a simple PointNet++~\cite{pointnet++} for feature extraction, with 3 layers of set abstraction and 3 layers of feature propagation.
Then, we design a flow prediction network $\mathcal{F}_f(\cdot)$ to predict the point cloud flow $f_{t-1 \rightarrow t} \in \mathbb{R}^{N \times 3}$ from frame $t-1$ to $t$, which contains the correspondence information between the two frames and represents the hand movement and object deformation:
\begin{equation} \label{eq:corr}
    f_{t-1 \rightarrow t} = \mathcal{F}_f(F_t, F_{t-1}, \mathcal{P}_t, \mathcal{P}_{t-1}).
\end{equation}




The flow prediction module consists of several MLPs and convolutional networks. After obtaining the point cloud flow $f_{t-1 \rightarrow t}$, we use another PointNet++ $I^\prime(\cdot)$ to extract the visual correspondence feature $F_t^f$ from $\mathcal{P}_{t-1}$ with $f_{t-1 \rightarrow t}$ added. $F_t^f$ and $F_t$ are then forwarded to a transformer fusion module to obtain the final visual feature $F_v$, corresponding to the current frame's point cloud $\mathcal{P}_t$. Specifically, the transformer fusion module first uses self-attention to encode the point cloud features. After applying positional embedding between the feature and its point cloud, a cross-attention module fuses the two features and outputs $F_v$. This fusion strategy considers both 3D static information of the current frame and the corresponding feature extracted from the flow. We will discuss the detailed architecture of the flow prediction module, the importance of it, and flow prediction results in the supplementary material.

\textbf{Object Decoder.}
The \textit{object decoder} processes the final visual feature $F_v$ in two steps: \textit{feature scattering \& sampling} and \textit{SDF decoding}, to predict SDF values and reconstruct the object mesh using the Marching Cubes algorithm \cite{marching_cubes}. Following ConvOccNet~\cite{convocc}, we scatter features into a volume of resolution $D$, process them with a 3D-UNet, and predict signed distances using a 5-layer MLP.

\textbf{Hand Decoder.}
For hand reconstruction, we use the parametric MANO model \cite{mano}, with $\beta \in \mathbb{R}^{10}$ for shape and $\theta \in \mathbb{R}^{51}$ for pose. Based on $F_v$ and the previous frame's point cloud $\mathcal{P}_{t-1}$, we predict hand joint positions $\mathcal{J}_h \in \mathbb{R}^{21 \times 3}$ using a voting mechanism inspired by PVN3D~\cite{pvn3d}. This involves predicting a translation offset $\mathcal{O}_t \in \mathbb{R}^{21 \times 3}$ and a probability matrix $T_{t-1} \in \mathbb{R}^{N \times 21}$ to compute:
\begin{equation}
    \mathcal{J}_h = \mathcal{O}_t + \mathcal{P}_{t-1} \times T_{t-1}.
\end{equation}
Joint poses $\theta$ are then estimated via inverse kinematics \cite{zhou2020monocular}, and the hand mesh is reconstructed using the MANO layer.


\textbf{Contact Constraint. }\label{para:c_constrain}
In order to obtain a relatively reasonable contact state of the reconstructed hand and object, we use a contact constraint to better optimize the prediction of the SDF of every sample position. Assuming we have the contact information $c_x \in \{0, 1 \}$ of a sample position $x \in \mathcal{X}$, then if $c_x = 1$, it indicates that at this position, the object is in contact with the hand, meaning its signed distance value $s_x$ should be $0$, and otherwise not equal to $0$. Therefore, the contact constraint should be as follows:
\begin{equation}
    \mathcal{L}_C = \sum_{x \in \mathcal{X}}s_x \cdot \mathds{1}_{c_x = 1}.
\end{equation}
If the tactile readings $\mathcal{M}$ can be obtained by haptic sensors, we can locate regional tactile arrays $\mathcal{M}^j$ through the hand reconstruction result, and the contact state $c_x$ of sample points whose distances are less than $l_c$ from those active tactile sensors will be set to 1, otherwise 0. In the absence of access to tactile signals, we can also use a simple network to predict contact states, which consists of a PointNet-based classifier with the input of sample positions and the final visual feature $F_v$. We will discuss the different effects of tactile readings and network prediction for acquiring interaction states in Sec. \ref{sec:ablation}.

\textbf{Loss Function.}
The network is trained end-to-end with four loss terms:
\begin{equation}
    \mathcal{L} = \lambda_f \mathcal{L}_{flow} + \lambda_S \mathcal{L}_{SDF} + \lambda_H \mathcal{L}_{Hand} + \lambda_C \mathcal{L}_{C}.
\end{equation}
Flow loss $\mathcal{L}_{flow}$ uses Chamfer distance between forward- and backward-shifted point clouds. SDF loss $\mathcal{L}_{SDF} = |s - s^*|$ compares predicted and ground truth SDF values. Hand joint loss $\mathcal{L}_{Hand} = \left\lVert \mathcal{J}_h - \mathcal{J}_h^* \right\rVert_2^2$ penalizes deviations from ground truth keypoints. The contact constraint $\mathcal{L}_C$ ensures accurate SDF predictions based on contact states.

\subsection{Force-aware Hand-pose Optimization}
The visual dynamic tracking net can give decent outputs of hand-object reconstruction. However, even though the contact constraint module has taken contact states into account for object reconstruction, the hand pose still needs refinement, for further improving the contact details. Here, we introduce the force-aware optimization for hand pose.

Given the contact energy defined in Sec. \ref{sec:cpf}, we optimize the predicted pose $\theta$ from the hand decoder, with respect to the reconstructed object mesh, to obtain a better hand mesh and contact state.
Based on Eq. \ref{eq:energy}, we use the ball-query method to find the corresponding object vertices for each hand region with a radius $R$, and the point pair will be set up between them and the hand keypoints $\mathcal{J}_h$.

Besides, For a given joint $j$, we ensure it remains in reasonable poses by penalizing rotations $\mathcal{R}_j$ that are near twisted directions $\mathcal{R}_t$ or if any angle exceeds $\pi / 2$, using the $L_2$ loss. Additionally, we constrain the optimized hand pose $^*\theta$ to stay close to the original prediction:
\begin{equation}
    \begin{aligned}
    \mathcal{L}_r & = \left\lVert \mathcal{R}_j \cdot \mathcal{R}_t \right\rVert_2^2 
     + \left\lVert \max(|\mathcal{R}_j| - \frac{\pi}{2}, 0) \right\rVert_2^2, \\
     \mathcal{L}_o & = \left\lVert ^*\theta - \theta \right\rVert_2^2.
     \end{aligned}
\end{equation}

Finally, the optimization target can be demonstrated as:
\begin{equation}
    ^*\theta = \mathop{\arg\min}_{\theta} (E + \mathcal{L}_r + \mathcal{L}_o).
\end{equation}

We use the gradient descent method and the Adam solver for optimization. By minimizing energy and loss, we can obtain a better interaction state between the hand and object, avoid severe penetration problems, and forbid the hand from abnormal poses. 

%% file: sec/5_dataset.tex
\begin{table*}[t!]
    \center
    \begin{tabular}{c|c|c|c|c|c}
        \toprule
        Metrics & IoU(\%)$\uparrow $ & CD(mm)$\downarrow$ & MPJPE(mm)$\downarrow$ & PD(mm)$\downarrow$ & CIoU(\%)$\uparrow$ \\ 
        \midrule
        \multicolumn{6}{c}{DexYCB} \\
        \midrule
        
        gSDF \cite{gsdf} (RGB) & 86.8 & 13.4 & 14.4 & \textbf{8.9} & 31.3 \\ 
        HOTrack \cite{hotrack} & 88.2 & 10.2 & 25.7 & 12.3 & 28.5 \\ 
        Zhang \etal\ \cite{zhang} & 88.1 & 10.4 & 15.3 & 10.6 & 30.5 \\ 
        Ours (w/o Force Opt.) & \textbf{90.1} & \textbf{9.6} & \textbf{13.2} & 9.9 & \textbf{35.4} \\ 

        \midrule

        \multicolumn{6}{c}{HOT} \\
        \midrule
        Zhang \etal\ \cite{zhang} & 77.9 & 16.8 & 15.6 & 11.9 & 26.8 \\ 
        ViTaM \cite{vitam} & 80.5 & 11.5 & 15.1 & 10.6 & 28.5 \\ 
        Ours (w/o Force Opt.) & \textbf{81.0} & \textbf{10.9} & 13.6 & 10.7 & 29.8 \\ 
        Ours (w. CPF \cite{cpf}) & * & * & 11.9 & 7.8 & 37.5 \\
        Ours (w. TOCH \cite{toch}) & * & * & 13.5 & 9.3 & 34.1 \\
        Ours (w. Force Opt.) & * & * & \textbf{11.3} & \textbf{7.3} & \textbf{40.3} \\

        \bottomrule

    \end{tabular}
    \caption{Quantitative results on DexYCB and HOT datasets for previous SOTA and our method. We only compare our results on rigid objects with the baselines. * indicates that adding hand-object refinement methods doesn't impact the metrics on reconstructed objects. $\uparrow / \downarrow$ indicates higher scores/lower scores are better. Force Opt. is short for force-aware optimization.}
    \label{tab:quan}
    \vspace{-0.3cm}
\end{table*}

\section{Hand-Object Tactile Dataset, HOT Dataset} \label{sec:sim}
Obtaining ground truth for object deformation in real-world scenarios is challenging, and prior works \cite{dexycb,oakink} on hand-object interaction primarily focus on rigid or articulated objects. To benchmark our method's performance on deformable objects, we introduce the \textbf{H}and-\textbf{O}bject-\textbf{T}actile (\textbf{HOT}) dataset, featuring depth images, force sensor readings, and SDF ground truths generated using the ZeMa simulator \cite{zema}. ZeMa employs FEM for intersection-free, high-accuracy contact modeling and deformation.

\subsection{Data Acquisition}
An example of data collection in the simulation environment is shown in Fig. \ref{fig:sim}.b. To ensure stable grasp with multiple contacts, we use DiPGrasp \cite{dipgrasp}, a fast grasp planner. After obtaining a feasible grasp pose, we position the hand near the target, add slight rotational perturbations, and set the hand fully open. It then moves toward the target and performs the grasp, generating a sequence of depth images, point clouds, and tactile arrays $\{\mathcal{P}_t, \mathcal{M}_t\}_{t=1}^n$ per frame.

\subsection{Dataset Statistics}
The \textbf{HOT} dataset includes objects from the YCB repository \cite{ycb}, with 5-10 objects from the \textbf{Bottle} and \textbf{Box} categories. We also add 5 \textbf{Sponges}, 5 \textbf{Plasticines}, and 5 \textbf{Stuffed Toys} to evaluate large deformations. Object properties are defined using density $\rho$, Young's modulus $E$, and Poisson's ratio $\nu$. For \textbf{Bottle} and \textbf{Box}, $\rho = 10^3 kg/m^3$, $E = 1.271$GPa, $\nu = 0.28$; for \textbf{Sponges}, \textbf{Plasticines}, and \textbf{Stuffed Toys}, $\rho = 30 kg/m^3$, $E = 0.1$MPa, $\nu = 0.38$. \textbf{Plasticine} has a yield stress $s_y = 200$pa.

Each object has 20 trajectories (15 training, 2 testing, 3 validation), totaling 600 sequences of 50-100 frames. Each sequence includes point clouds from 8 camera views.

%% file: sec/6_experiments.tex
\section{Experiments}

\begin{figure*}[t!]
    \centering
    \includegraphics[width=0.8\linewidth]{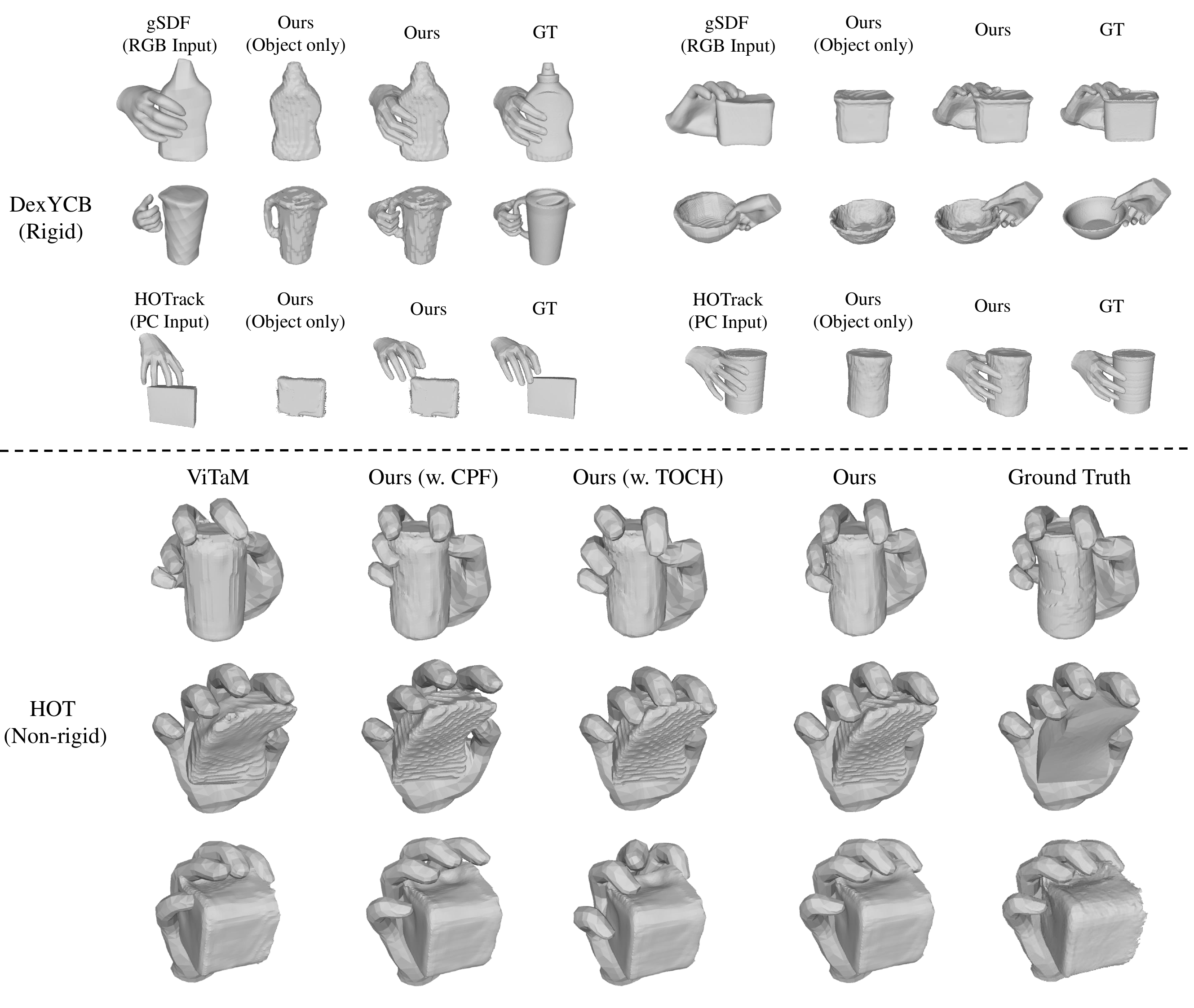}
    \caption{Qualitative results on both DexYCB and HOT datasets. The upper shows our better performances on rigid objects compared to the baselines, and the lower demonstrates the effectiveness of model for dynamic tracking of deformable objects and hands. }
    \label{fig:qual}
    \vspace{-0.5cm}
\end{figure*}

\subsection{Implementation Details}
We use Farthest Point Sampling to downsample the input point cloud size to $N = 1024$ and sample $1 \times 10^6$ positions in space, including $2 \times 10^5$ surface points. During training, we subsample $M = 2048$ for SDF prediction. The volume scattering feature resolution $D=64$. The distance threshold in contact constraint $l_c = 3$mm. Loss weights are $\lambda_f = 0.01$, $\lambda_S = 0.5$, $\lambda_H = 1$, and $\lambda_C = 0.05$.

The visual dynamic tracking net is trained on the entire dataset with a batch size of 6, learning rate $1e-4$, and Adam optimizer for 100 epochs. Fine-tuning per object category uses a batch size of 4, learning rate $5e-5$, and 50-100 epochs. Training takes 15 hours on an Nvidia A40 GPU.

For force-aware optimization, the barrier function threshold $\hat{l} = 2$mm and the ball-query radius $R=5$mm. The energy is minimized for 100 iterations per frame with a learning rate of $2e-3$, taking $3.5 \pm 0.5$ seconds per frame.

\subsection{Datasets}\label{sec:data}
We use \textbf{DexYCB} for rigid object testing and the \textbf{HOT} dataset for deformable object interactions. DexYCB validates our method against baselines, while HOT evaluates deformation tracking and force-aware contact modeling.

\subsection{Metrics}\label{sec:metrics}
\noindent\textbf{Intersection over Union (IoU)} measures mesh overlap between predictions and ground truth.

\noindent\textbf{Chamfer Distance (CD)} evaluates object vertex reconstruction accuracy.

\noindent\textbf{Mean Per Joint Position Error (MPJPE)} assesses hand joint tracking precision.

\noindent\textbf{Penetration Depth (PD)} quantifies hand-object interaction plausibility by measuring maximum hand penetration into the object.

\noindent\textbf{Hand-Object Contact Mask IoU (CIoU)} validates contact recovery by comparing contact masks (defined as distances $< 3$mm) between optimized results and ground truth.

\subsection{Results}\label{sec:res}

For hand-object tracking evaluation, we compare our method with state-of-the-art approaches: gSDF \cite{gsdf}, HOTrack \cite{hotrack}, Zhang \etal\ \cite{zhang}, ViTaM \cite{vitam}, CPF \cite{cpf}, and TOCH \cite{toch}. We include gSDF which uses RGB images only for metrics references, and \cite{zhang} uses depths for SDF-based object tracking. HOTrack predicts object poses from segmented point clouds, and ViTaM, similar to our method, takes unsegmented point clouds and tactile arrays as input but uses WNF representation, making it a strong baseline. We compare gSDF and HOTrack on DexYCB for rigid objects without tactile input, and ViTaM on the HOT dataset due to its tactile requirement. CPF and TOCH are used to benchmark our force-based optimization. We will discuss how to use force-aware optimization empirically with a fixed force setting in Sec. \ref{sec:ablation}.

Quantitative results in Tab. \ref{tab:quan} show our method outperforms baselines on DexYCB in IoU and Chamfer distance, demonstrating superior object tracking and reconstruction. Our higher MPJPE score indicates better hand-tracking performance. On the HOT dataset, our visual dynamic tracking net matches ViTaM's performance, while force-based optimization significantly improves MPJPE, PD, and CIoU, confirming its effectiveness. This is largely due to our method's ability to obtain accurate contact information through tactile arrays, unlike previous methods that rely on network-based contact modeling.

\begin{figure*}
    \centering
    \includegraphics[width=1\linewidth]{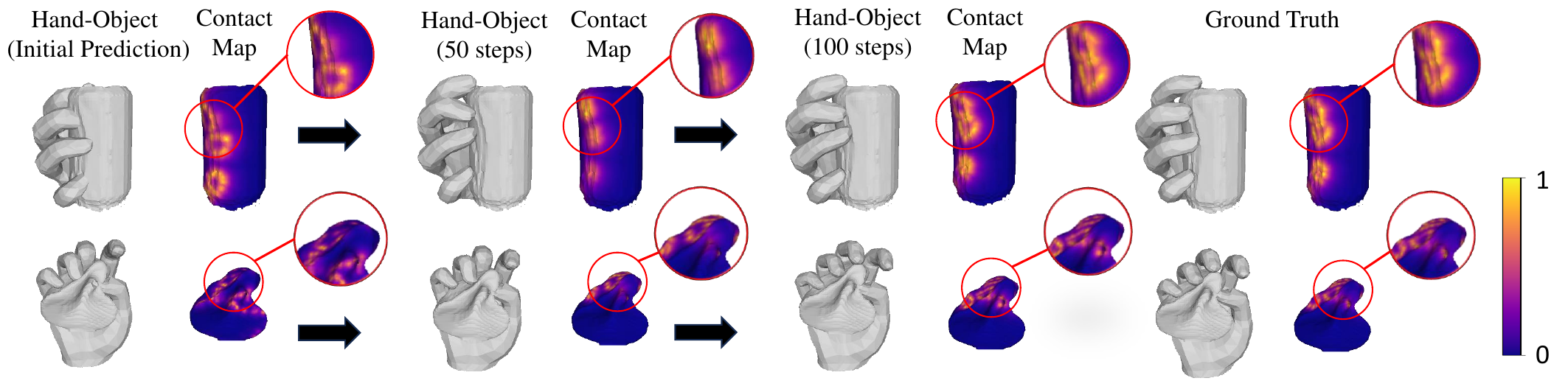}
    \caption{Qualitative results with force-aware optimization. As iteration steps rise, the penetration problem decreases, and the contact map is closer to the ground truth, making the contact more reasonable.}
    \label{fig:qual_geo}
    \vspace{-0.5cm}
\end{figure*}

Figure \ref{fig:qual} shows the qualitative results of our method. Our tracking and reconstruction outperform state-of-the-art methods in the DexYCB dataset, and our force-aware optimization process shows excellent performance on deformable objects in the HOT dataset in comparison with previous hand-object optimization baselines. Our method accurately tracks hand movements and object deformations, thanks to the fusion of flow features and original point cloud features using a transformer, which incorporates both the 3D information of the current frame and correspondence features from the previous frame.
Figure \ref{fig:qual_geo} visualizes the force-aware hand-pose optimization. With force-aware contact representation DF-Field, the optimization can solve most of the penetration problems and refine the contact map based on collected tactile information.

\subsection{Ablation Study}\label{sec:ablation}

We conduct the ablation study on (1) methods for obtaining contact states for the contact constraint (Sec. \ref{para:c_constrain}) and (2) definitions of force representation (heuristic vs. tactile-aware). Additional discussions on point pair establishment, flow prediction, and tactile array integration in the visual dynamic tracking net are in the supplementary material.

\noindent\textbf{Contact State \& Effect of Contact Constraint. }
We evaluate the impact of the contact constraint by testing the visual dynamic tracking net without it. Additionally, we compare ground truth contact states with predictions from a PointNet-based network when there is a lack of tactile hardware support. Results in Tab. \ref{tab:abl_contact} show that the contact constraint significantly improves object reconstruction and hand-object interaction prediction. The PointNet-based network effectively predicts interaction states from visual inputs, while tactile readings match ground truth accuracy, greatly enhancing object geometry reconstruction.


\begin{table}
    \center
    \begin{tabular}{c|c|c|c|c}
        \toprule
        Metrics & IoU$\uparrow$ & CD$\downarrow$ & CIoU$\uparrow$ & MPJPE$\downarrow$ \\ 
        \midrule

        w/o CC & 75.9 & 12.8 & 25.3 & 12.0  \\ 
        CC. (GT) & \textbf{81.2} & 11.2 & 29.2 & \textbf{11.9} \\ 
        CC. (Tactile) & 81.0 & \textbf{10.9} & \textbf{29.8} & \textbf{11.9} \\ 
        CC. (PointNet) & 78.3 & 12.1 & 27.6 & 12.1 \\ 

        \bottomrule

    \end{tabular}
    \caption{Quantitative results for different contact information acquirement. CC. indicates Contact Constraint, along with approaches for acquiring contact information.}
    \label{tab:abl_contact}
    \vspace{-0.3cm}
\end{table}

\begin{table}
    \center
    \begin{tabular}{c|c|c|c}
        \toprule
        Metrics & MPJPE$\downarrow$ & PD$\downarrow$ & CIoU$\uparrow$ \\ 
        \midrule
        \multicolumn{4}{c}{DexYCB} \\
        \midrule

        w/o Force Opt. & 13.2 & 9.9 & 35.4 \\ 
        $\mathcal{M}$ fixed Force Opt. & \textbf{12.3} & \textbf{8.5} & \textbf{39.7} \\

        \midrule
        \multicolumn{4}{c}{HOT} \\
        \midrule

        w/o Force Opt. & 13.6 & 10.7 & 29.8 \\ 
        $\mathcal{M}$ fixed Force Opt. & 12.9 & 8.5 & 36.8 \\ 
        Force Opt. & \textbf{11.3} & \textbf{7.3} & \textbf{40.3} \\

        \bottomrule

    \end{tabular}
    \caption{Quantitative results for different force representations in force-aware hand-pose optimization.}
    \label{tab:abl_k}
    \vspace{-0.6cm}
\end{table}

\noindent\textbf{Different Force Representations.}
We explore the effect of force representations on contact modeling. For DexYCB (no tactile data), we set $\mathcal{M}_j = 0.5$ for regions near objects ($l=3mm$) and compare results with the visual dynamic tracking net. For HOT, we test the optimization with $\mathcal{M}_j = 0.5$ to evaluate the impact of removing tactile information.

Results in Tab. \ref{tab:abl_k} demonstrate that even with fixed $\mathcal{M}$, DexYCB performance improves, validating the effective design of DF-Field. On HOT, tactile information significantly enhances penetration resolution and contact recovery, as distributed tactile arrays provide accurate interaction forces, outperforming empirical force settings.

\subsection{Real-world Experiments}
In order to validate the effectiveness of our method, we conduct real-world experiments with our network trained on simulation data. We bought a ViTaM glove \cite{vitam} from its authors to capture distributed tactile sensory data and a Photoneo MotionCam M+ high-precision depth camera to record the point cloud series. We can see an example of a reconstruction result on an unseen stuffed toy in Fig. \ref{fig:real}. We will demonstrate our real-world setup and more series of tracking results in supplementary material.

\begin{figure}
    \centering
    \includegraphics[width=0.9\linewidth]{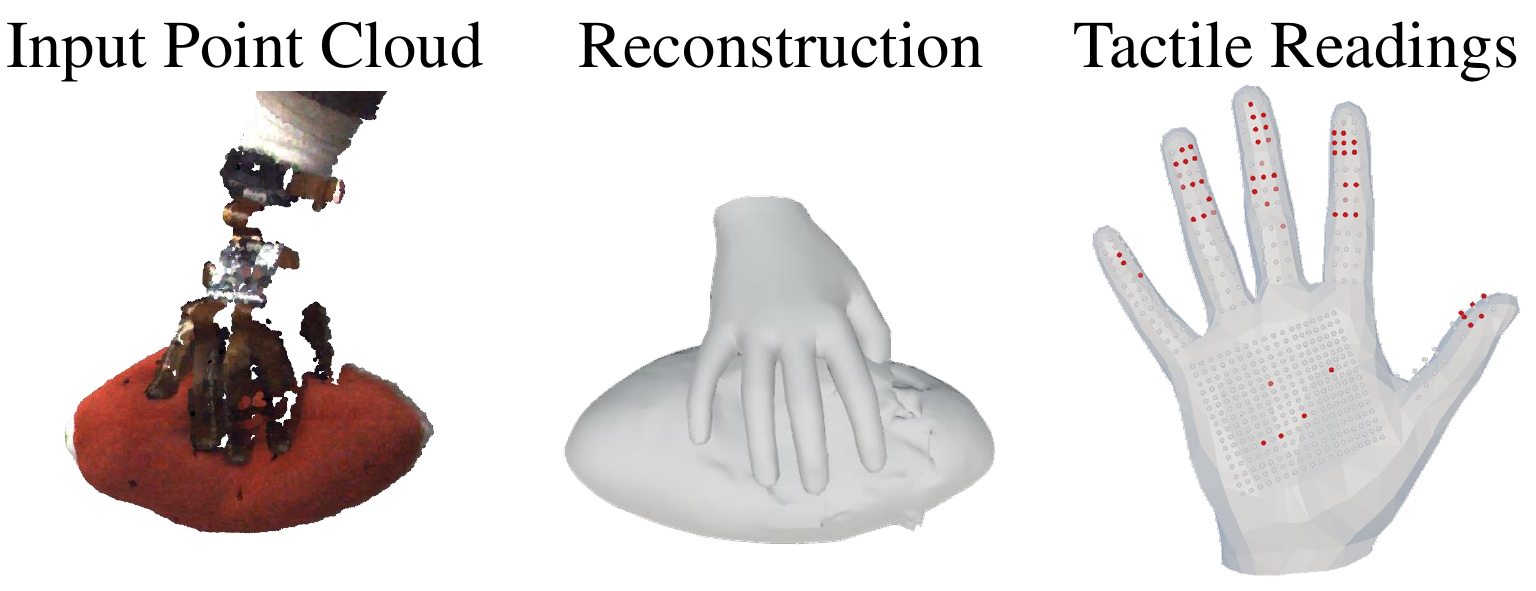}
    \caption{Real-world reconstruction example of an unseen stuffed toy. The overall shape and deformation of the toy have been well reconstructed.}
    \label{fig:real}
    \vspace{-0.5cm}
\end{figure}

%% file: sec/7_conclusion.tex
\vspace{-0.2cm}
\section{Conclusion and Future Works}
In this work, we introduce ViTaM-D, a dynamic hand-object interaction reconstruction framework that integrates distributed tactile sensing with visual perception. Featuring DF-Field and force-aware optimization, our approach effectively captures fine contact details and object deformation. We also present the HOT Dataset for benchmarking deformable object manipulation. Evaluations show that ViTaM-D outperforms state-of-the-art methods on both rigid and deformable objects. Future works include applying ViTaM-D to data collection for robotic imitation learning, dexterous manipulation, and human-robot interaction.